\documentclass[sigconf]{acmart}
\usepackage{multirow}
\usepackage{graphics}
\usepackage{epstopdf}

\AtBeginDocument{%
	\providecommand\BibTeX{{%
			\normalfont B\kern-0.5em{\scshape i\kern-0.25em b}\kern-0.8em\TeX}}}

\copyrightyear{2022}
\acmYear{2022}
\setcopyright{acmlicensed}

\acmConference[MM '22] {Proceedings of the 30th ACM International Conference on Multimedia}{October 10--14, 2022}{Lisboa, Portugal.}
\acmBooktitle{Proceedings of the 30th ACM International Conference on Multimedia (MM '22), Oct. 10--14, 2022, Lisboa, Portugal}
\acmPrice{15.00}
\acmISBN{978-1-4503-9203-7/22/10}
\acmDOI{10.1145/XXXXXX.XXXXXX}

\usepackage{multirow}
\usepackage{amsmath}
\usepackage{multirow}
\usepackage[T1]{fontenc} 
\usepackage{microtype} 
\usepackage{diagbox}
\usepackage[english]{babel} 
\usepackage{lettrine} 

\usepackage{enumitem} 
\setlist[itemize]{noitemsep} 
\usepackage{hyperref} 
\usepackage{algorithm}
\usepackage{algorithmic}
\usepackage{marvosym}
\allowdisplaybreaks
\AtBeginDocument{%
  \providecommand\BibTeX{{%
    Bib\TeX}}}

\copyrightyear{2022}
\acmYear{2022}
\setcopyright{acmcopyright}
\acmConference[MM '22] {Proceedings of the 30th ACM International Conference on Multimedia}{October 10--14, 2022}{Lisboa, Portugal.}
\acmBooktitle{Proceedings of the 30th ACM International Conference on Multimedia (MM '22), Oct. 10--14, 2022, Lisboa, Portugal}
\acmPrice{15.00}
\acmISBN{978-1-4503-9203-7/22/10}
\acmDOI{10.1145/XXXXXX.XXXXXX}

\begin{document}

\title{Domain Camera Adaptation and Collaborative Multiple Feature Clustering for Unsupervised Person Re-ID}

\author{Yuanpeng Tu}
\authornote{Corresponding author.}
\affiliation{%
	\institution{Tongji University}
	\city{Shanghai}
	\country{China}}
\email{2030809@tongji.edu.cn}

\renewcommand{\shortauthors}{Tu, et al.}

\begin{abstract}
	Recently unsupervised person re-identification (re-ID) has drawn much attention due to its open-world scenario settings where limited annotated data is available. Existing supervised methods often fail to generalize well on unseen domains, while the unsupervised methods, mostly lack multi-granularity information and are prone to suffer from confirmation bias. In this paper, we aim at finding better feature representations on the unseen target domain from two aspects, 1) performing unsupervised domain adaptation on the labeled source domain and 2) mining potential similarities on the unlabeled target domain. Besides, a collaborative pseudo re-labeling strategy is proposed to alleviate the influence of confirmation bias. Firstly, a generative adversarial network is utilized to transfer images from the source domain to the target domain. Moreover, person identity preserving and identity mapping losses are introduced to improve the quality of generated images. Secondly, we propose a novel collaborative multiple feature clustering framework (CMFC) to learn the internal data structure of target domain, including global feature and partial feature branches. The global feature branch (GB) employs unsupervised clustering on the global feature of person images while the Partial feature branch (PB) mines similarities within different body regions. Finally, extensive experiments on two benchmark datasets show the competitive performance of our method under unsupervised person re-ID settings.
\end{abstract}


\maketitle

\section{Introduction}                  
\begin {figure}[t]
\centering
\includegraphics[width=0.75\linewidth]{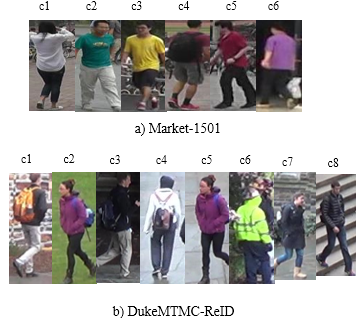}
\caption{Examples from Market-1501 and DukeMTMC-ReID dataset. Samples in the first and second rows are from different cameras in Market-1501 and DukeMTMC-ReID datasets, respectively. Images from different cameras show different styles. It is a very challenging task to identify the same person from different views.}
\label{fig:dataset}
\end{figure}

Person Re-Identification (re-ID) targets to retrieve images from a database collected by non-overlapping cameras. Existing re-ID methods \cite{zhong2017re,li2018harmonious} have achieved dramatic performances when training and testing on the same domain. Nevertheless, researchers have consistently shown that models trained on the source domain may have a significant performance drop when directly applied to the target domain since there exists a large domain gap between the source and target domain. As shown in Fig. \ref{fig:dataset}, since different datasets are often collected from different environments, images from different datasets often have large appearance variations in illumination and background clutter. Besides, in real-world applications, the training data on the target domain is usually unlabeled or partially labeled.

To tackle these issues, recently, considerable literature has grown up around the theme of unsupervised domain adaptation (UDA). Compared to generic unsupervised domain adaptation (UDA), the source and target domains in person re-ID have entirely different classes (i.e. person identities), which is studied as an open set domain adaptation problem. A few works \cite{deng2018image, zhu2017unpaired, wei2018person, zhong2018generalizing} have been proposed to deal with it, which mainly attempt to solve this problem by transferring the global style of the source domain to the target domain. However, intra-domain image variations also exist because of the disparities in cameras. Note that examples in each row of  Fig. \ref{fig:dataset} are sampled from different cameras in the same dataset and are distinct in the background. Thus, translating the global style of images may confuse the generative networks and impair the quality of generated images.

Apart from these methods, some techniques \cite{ge2020mutual, yu2019unsupervised} have been proposed to address the unsupervised person re-ID problem by pseudo label estimation. Features of unlabeled target training data are extracted from the model pre-trained on the labeled source dataset. Unsupervised clustering is applied to these features to generate pseudo labels, which are used for subsequent supervised training. The model training and unsupervised clustering are executed alternately. Clustering quality is critical to their performance. However, performing unsupervised clustering on the whole image may suffer from severe background clutter. Besides, the label generation lacks multi-granularity information and is prone to suffer from confirmation bias.

In this paper, we aim at finding better feature representations on the unseen target domain/dataset from two aspects, using labeled source training images and unlabeled target training images. First, considering the disparities of cameras within the source and target datasets, images from source camera sub-domains are transferred to target camera sub-domains while keeping the person identity. The generated images with specific person identity are thus utilized for supervised learning. Secondly, to mine the potential identity similarities on the target training set, a two-branch framework is proposed for similarity learning using multiple features to alleviate confirmation bias. Specifically, we propose a novel collaborative multiple feature clustering framework (CMFC) for learning representations on the target dataset: including a global feature guided training branch and a partial feature guided training branch. The global feature branch (GB) performs unsupervised clustering on global features and finetunes the network based on the cluster groups of images. The partial feature branch (PB) divides the person image into upper body region and lower body region and performs unsupervised clustering on different parts to obtain pseudo labels of person images. Pseudo labels are utilized to optimize the person re-ID models in a supervised manner. Both global features and partial features are combined in CMFC to learn the internal data distribution on the target dataset. Finally, a collaborative pseudo re-labeling strategy is proposed to alleviate the influence of confirmation bias. Finally, we summarize our main contributions as follows:
\begin{enumerate}
\item We propose a cross-domain camera style adaptation module to transfer images from the source dataset to the target dataset on the camera level, preserving the person identity. The transferred images are then fed to the models to obtain more discriminative features in a supervised manner.
\item We propose a collaborative multiple feature clustering framework (CMFC) to learn identity similarities on the target domain using multiple features and meanwhile alleviate the influence of confirmation bias. A global feature branch is employed to extract the global feature of pedestrian images and unsupervised clustering is performed on the global features to learn identity similarities. The partial feature branch divides the person images into different parts and the re-ID model is optimized by using the potential similarities in different parts. 
\item Through experimental results on two large-scale datasets, we demonstrate the effectiveness of our method and different components on unsupervised person re-ID.
\end{enumerate}

\section{Related work}\label{sec:relatedwork}               
In this section, since the proposed methods mainly focus on the unsupervised domain adaptation and unsupervised person re-ID methods, we briefly review several related works. 

\subsection{Supervised Person Re-identification}         
Extensive supervised methods have been developed on the widely used benchmarks \cite{zheng2015scalable,zheng2017unlabeled,ristani2016performance}, concentrating on discriminative feature representation learning \cite{li2018harmonious,bai2020deep}, deep metric learning \cite{yu2018hard,hu2015deep}, postprocessing procedures \cite{bai2017scalable,zhong2017re,cao2020progressive}, and other problems such as occlusion \cite{he2019foreground}, various image resolutions \cite{wang2018cascaded}. Although great progress has been observed, these supervised approaches may have a significant performance drop when applied to another unseen domain due to the existence of domain shift.

\subsection{Unsupervised Domain Adaptation}            
Unsupervised domain adaptation (UDA) aims to transfer the knowledge on a labeled source domain to the unlabeled target domain. Several unsupervised domain adaptation methods try to reduce the distribution discrepancy between source and target domain on either feature-level \cite{peng2016unsupervised,hu2015deep} or image-level \cite{deng2018image, wei2018person}. The former focuses on learning domain-invariant representations by aligning the feature statistics, such as the mean and covariance of source domain and target domain feature distributions \cite{sun2015return}, or the adversarial training approach. The latter learns to transform samples in the pixel space from the source domain to the target domain using generative adversarial networks \cite{bousmalis2017unsupervised, hoffman2018cycada}. Nevertheless, most of the existing unsupervised domain adaptation methods assume that source and target domains share the same set of classes, while in person re-ID task the source and target domains have entirely different classes. Thus, these methods cannot be directly utilized for unsupervised domain adaptation in person re-ID.

\subsection{Unsupervised Person Re-identification}   
Unsupervised person re-ID methods are proposed to utilize the unlabeled target data and large-scale labeled samples. Among them, some works try to address this problem with domain adaptation on feature or image levels. In \cite{yang2020part}, PPAN is proposed to enforce feature alignment across domains. Peng et al. \cite{peng2016unsupervised} propose to learn based on asymmetric multi-task dictionary learning. Other works \cite{deng2018image, wei2018person} attempt to transfer the source domain images to target domain styles using generative adversarial networks (GAN). Deng et al. \cite{deng2018image} introduce a similarity preserving cycle consistent generative adversarial network (SPGAN) to translate images. However, intra-domain image variations still exist because of the distribution discrepancy at the camera level. HHL \cite{zhong2018generalizing} considers the intra-domain image style variations caused by different camera configuration. M2MGAN \cite{liang2018m2m} takes multiple source and target sub-domains into consideration. Different from these methods, our cross-domain camera style adaptation module explicitly considers the camera-level disparities and transforms images from the source domain to different cameras in the target domain. Moreover, we introduce person identity preserve loss constraint and identity mapping loss constraint to improve the quality of the generated images. Finally, considering the confirmation bias of label generation, a collaborative pseudo re-labeling strategy is proposed.

\begin {figure*}[t]

\centering
\includegraphics[width=1.0\linewidth]{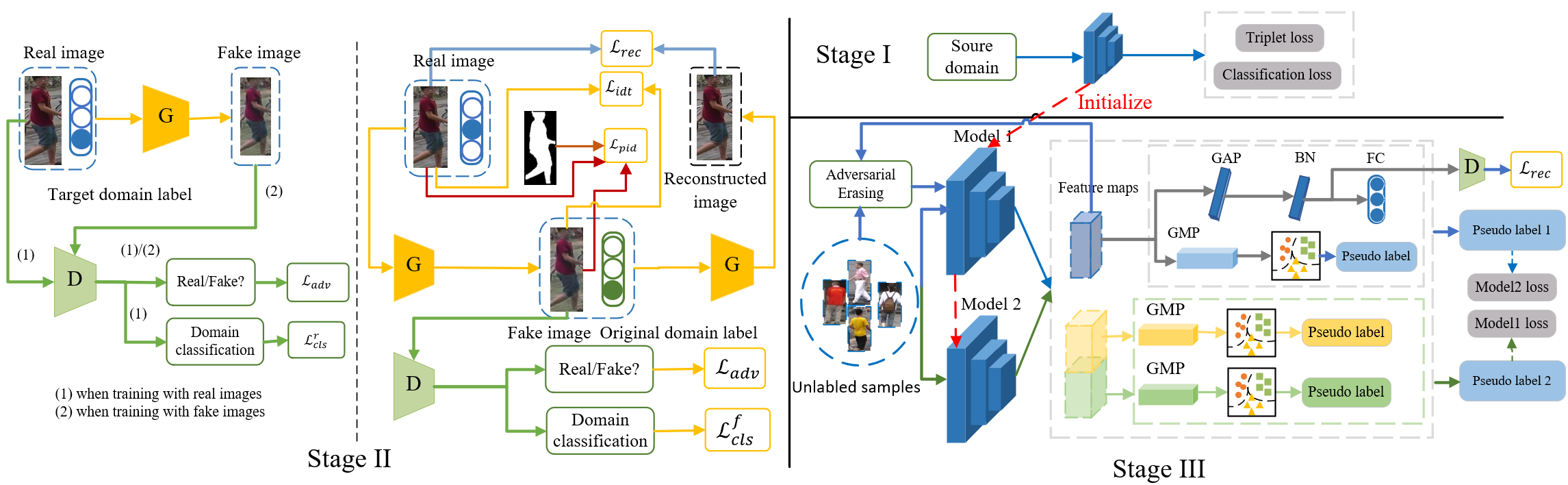}

\caption{The overall framework of the proposed method. Specifically, the proposed method trains the re-ID model in three stages. The first stage learns a pre-trained re-ID baseline model on a labeled source training dataset. Based on the baseline, the second stage trains the model with images transferred from the source dataset to the target dataset, performing multi-domain image translation. In the third stage, we conduct unsupervised clustering and cross-finetune two models with pseudo labels from each other on the target training set alternately to learn the discriminative re-ID model.}

\label{fig:framework}
\end{figure*}

Beyond the above methods, some approaches focus on label estimation in the target domain. Fan et al. \cite{fan2018unsupervised} propose an unsupervised re-ID approach for iteratively applying k-means clustering. Yang et al. \cite{yang2019leveraging} generate labeled virtual data from the target dataset and propose collaborative filtering on unlabeled data. A Self-similarity Grouping (SSG) approach \cite{fu2019self} iteratively conducts grouping and re-ID model training in a self-paced manner. A self-training method with progressive augmentation \cite{zhang2019self} jointly captures the local structure and global data distribution. Soft multi-label learning \cite{yu2019unsupervised} mines the soft label information from a reference set for unsupervised learning. Inspired by these methods, we propose a collaborative multiple feature clustering framework (CMFC) to learn identity similarities using multiple features on the target domain. CMFC is a two-branch framework, including global and partial feature branches, which improves the accuracy of internal data similarity learning using multiple features.

\section{Proposed method}\label{sec:method}      
\textbf{\emph{Problem definition.}} 
For unsupervised person re-ID, we have a labeled dataset $\left\{ X_{S},Y_{S} \right\}$ from source domain, which contains N$_{S}$ person images. Each image x$_{s}^{i}$ corresponds to a label y$_{s}^{i}$,
where $ y_{s}^{i}\in \left\{1,\ 2,\ldots,\ M_{S} \right\}$.
M$_{S}$ is the number of identities in the source dataset. We also have an unlabeled dataset $\left\{ X_{T} \right\}$ from target domain, containing N$_{T}$ unlabeled images. Note that the identity of the target image is unknown while the camera label is available, which conforms to real-world settings. C$_{S}$  and C$_{T}$ denote the number of cameras in the source dataset and target dataset, respectively. 

\textbf{\emph{Motivation.}} 
The goal of this paper is to learn the discriminative embeddings of the target dataset by both leveraging the knowledge of the source dataset and mining internal similarities on the target dataset. Thus, in Section \ref{subsec:gan}, we perform cross-domain camera style adaptation to transfer images from the source domain to the target domain. In this way, the generated images with target domain styles can be trained in a supervised manner to improve the discriminative ability of re-ID model. Besides, in Section \ref{subsec:cluster}, we further explore the similarities on the target dataset and finetune the re-ID model with positive pairs and negative pairs to obtain discriminative feature representation on the target dataset.

\subsection{Supervised pre-training}\label{subsec:pretrain}    
To learn feature embeddings of the source domain, we train on the source dataset and denote the obtained model as \textbf{\emph{baseline}}. We use ResNet50 \cite{he2016deep} as backbone network. Given the labeled images in a training batch, we train the baseline model with cross-entropy loss and batch-hard triplet loss \cite{hermans2017defense} simultaneously. The cross-entropy loss is employed with the output of the FC layer by treating the training process as a classification task. Besides, label smoothing \cite{szegedy2016rethinking} is used to avoid overfitting. Specifically, the cross-entropy loss with label smoothing can be formulated as:
\begin{equation}
	\mathcal{L}_{cross}=-\frac{1}{n_s}\sum_{i=1}^{n_s}{\sum_{k=1}^{M_S}{q_i(k)\log p_i(k)}} 
\end{equation}

\begin{equation}
	q_i(k)= \left\{
	\begin{aligned}
		1-\varepsilon+ \frac{\varepsilon }{M_S}  &    &    k=y  \\ 
		\frac{\varepsilon }{M_S}      &  &                 k\neq y
	\end{aligned}
	\right.
\end{equation}
where n$_{s}$ is the number of images in a batch, $ p_{i}(k) \in \left[0,\ 1 \right] $
is the predicted probability belonging to class k. q$_{i}$(k) is the label distribution and y is the ground truth class label. $\varepsilon$ is a small perturbation term.
The triplet loss is employed to enhance intra-class compactness and inter-class separability, which can be written as,
\begin{equation}
	\mathcal{L}_{tri}=\sum_{x_a,\ x_p,\ x_n\ }\left[m+D_{x_a,\ {\ x}_p}-D_{x_a,\ x_n}\right]_+
\end{equation}
where x$_{a}$ denotes the anchor. x$_{p}$ and x$_{n}$ represent the hardest positive sample and the hardest negative sample in the same batch respectively. \emph{m} is a margin hyperparameter and \emph{D$(\bullet)$} is the Euclidean distance between two features. Thus, the overall loss function is written as follows
\begin{equation}
	\mathcal{L}_{baseline}=\mathcal{L}_{cross}+ \lambda_{t}\mathcal{L}_{tri}
\end{equation}
where $\lambda_t$ can be set to 1 in the experiments for simplicity.

\subsection{Cross Domain Camera Style Adaptation}\label{subsec:gan}      
To leverage the knowledge of the source dataset, we try to reduce the distribution discrepancy between the source and target dataset at the image level. Instead of transferring the global style of images, we perform image-image translation by viewing each camera as an individual domain. 

Given a dataset X$_{S}$ from the source domain and a dataset X$_{T}$ from the target domain where the camera labels are available, our goal is to train a single generator \emph{G} that learns the mappings among multiple domains. In this way, a given labeled image $\left( x_{s}^{i},y_{s}^{i} \right)$ with camera label c$_{s}^{i}$ is transferred to another camera style c$_{t}$ of target domain, while preserving the identity information during the translation. The generated images can then be used to train the person re-ID model. To learn the style mapping between source and target datasets, we employ StarGAN \cite{choi2018stargan} and further introduce two loss items during the image translation training procedure to improve the quality of the transferred images.


StarGAN trains a single generator \emph{G} to translate an input image \emph{x} into an output image \emph{y} conditioned on the target domain label \emph{c}, $G\left(x,\ c\right) \rightarrow y$. In addition, to distinguish real training examples and generated samples from \emph{G}, an auxiliary classifier is introduced to the discriminator \emph{D} to control multiple domains. Thus, the discriminator produces probability distributions over both sources and domain labels, $ D:x\rightarrow\left\{D_{src}\left(x\right),\ \ D_{cls}\left(x\right)\right\} $. The network architecture of StarGAN is described in Section \ref{subsec:implement details}.

\textbf{\emph{Adversarial Loss.}} 
Adversarial loss is adopted to make the generated images indistinguishable from real images. The generator \emph{G} tries to minimize the loss while the discriminator tries to maximize it.
\begin{equation}
	\mathcal{L}_{adv}=E_x\left[log{D_{src}\left(x\right)}\right]+E_{\ x,\\ c}[log{{(1-D}_{src}(G(x,c)))}
\end{equation}

\textbf{\emph{Domain Classification Loss.}} 
To distinguish the domain labels of a real/fake image, an auxiliary classifier is added on top of \emph{D} and cross-entropy loss is utilized to optimize both \emph{G} and \emph{D}. Specifically, the domain classification loss of real images is used to optimizer \emph{D} and the domain classification loss of fake images is used to optimize \emph{G}, \emph{that is},
\begin{equation}
	\mathcal{L}_{cls}^r=E_{x,c^\prime}\left[-logD_{cls}(c^\prime|x)\right]
\end{equation}

\begin{equation}
	\mathcal{L}_{cls}^f=E_{x,c}\left[-log{D_{cls}\left(c|G(x,c)\right)}\right]
\end{equation}
where \emph{c$^{\prime}$} is the original domain label of real image \emph{x}.

\textbf{\emph{Reconstruction Loss.}} 
To preserve the content of input images while changing only the domain-related style of images, reconstruction loss is used to formulate forward cycle consistency.
\begin{equation}
	\mathcal{L}_{rec}=E_{x,c,c^\prime\ }[{\parallel x-G(G\left(x,\ c\right),\ c^\prime)\parallel}_1]
\end{equation}

\textbf{\emph{Identity mapping Loss.}} 
Apart from adversarial loss, domain classification loss, and reconstruction loss, we introduce identity mapping loss to regularize the generator to be the identity matrix on samples from the target domain. SPGAN \cite{deng2018image} uses this loss to preserve the color composition between the input and output during translating images from the source dataset to the target dataset. The identity mapping loss is written as
\begin{equation}
	\mathcal{L}_{idt}=E_{x,c\ }{\parallel G\left(x,\ c\right)-x\parallel}_1
\end{equation}

\textbf{\emph{Person Identity Preserve Loss.}}
To utilize the transferred images to supervised person re-ID model training, it is important to preserve the identity of person images while changing the style of images. We introduce the person identity preserve loss by evaluating the variations in the person foreground before and after person transfer. However, the common identity preserve loss form is prone to suffer from perturbation in the foreground masks. To enhance the robustness of $G$, a consistency regularization item is introduced by constraining the output of the original image and the enhanced image to be consistent, which is shown as follows,
\begin{equation}
	\begin{split}
		\mathcal{L}_{pid}=E_{x,c\ }{\parallel G\left(x,\ c\right)\odot M(x)-x\odot M(x)\parallel}_2 \\
		+ E_{x,x^{a},c\ }{\parallel G\left(x,\ c\right)\odot M(x)-G(x^{a},c)\odot M(x^{a})\parallel}_2
	\end{split}
\end{equation}
where \emph{M(x)} and $x^a$ represents the foreground mask and augmented images of \emph{x}. In this paper, we use SOLO \cite{wang2019solo} instance segmentation algorithm to extract the foreground mask.

\textbf{\emph{Overall objective function.}}
Stage II in Fig. \ref{fig:framework} shows the overview of the training process of the proposed cross-domain camera style adaptation module. Specifically, the objective functions to optimize \emph{G} and \emph{D} are written, respectively, as
\begin{equation}
	\begin{split}
		\mathcal{L}_D=-\mathcal{L}_{adv}+\lambda_{cls}\mathcal{L}_{cls}^r   \\
		\mathcal{L}_G=\mathcal{L}_{adv}+\lambda_{cls}\mathcal{L}_{cls}^f+\lambda_{rec}\mathcal{L}_{rec}+{\lambda_{idt}\mathcal{L}}_{idt}+{\lambda_{pid}\mathcal{L}}_{pid}
	\end{split}
\end{equation}
where $\lambda_{cls}$, $\lambda_{rec}$, $\lambda_{idt}$, $\lambda_{pid}$ are hyper-parameters that control the relative importance of domain classification loss, reconstruction loss, identity mapping loss, and person identity preserve loss, respectively. Empirically, we use $\lambda_{cls}$=1, $\lambda_{idt}$=1, $\lambda_{rec}$=10 and $\lambda_{pid}$=10 in our experiments.

\subsection{Collaborative Multiple feature clustering framework (CMFC)}\label{subsec:cluster}     
In addition to performing unsupervised domain adaptation, mining potential label information (identity similarity) on the target domain is essential to unsupervised person re-ID. In this section, we introduce a collaborative multiple feature clustering framework (a two-branch network) based on the global features and partial features to mine the similarities of target images to train re-ID model. 

\textbf{\emph{Two-branch re-ID model.}} 
As illustrated in Stage III of Fig. \ref{fig:framework}, the two-branch network shares the same backbone with the baseline model. Before training on the target domain, the model is trained on the source dataset and transferred images described in Section \ref{subsec:gan}, learning useful representations of person images. However, the model is still not discriminative in the target domain. Afterward, two two-branch networks are trained parallelly with the pseudo labels from each other to alleviate confirmation bias. Our two-branch network includes two branches: the global feature branch and the partial feature branch. 

\textbf{\emph{Global feature branch.}}
Given an unlabeled image x$_{t}^{i}$, we first feed it into the pre-trained model for feature extraction. The feature map of image x$_{t}^{i}$ denotes as $F_{t}^{i}\in\mathbb{R}^{H\times W\times C}$ (the output of Resnet50 layer5). Next, we employ the global max pooling (GMP) on the feature map to obtain feature vector f$_{t}^{i}$. For every image in target dataset, we extract the feature vector to form a feature vector set $\left\{f_t^1,\ ...,\ f_t^k,...,f_t^{N_t}\right\} $. Based on the vector set, an unsupervised clustering algorithm is utilized to divide the target dataset into different groups. In this paper, we use DBSCAN \cite{ester1996density} algorithm to perform unsupervised clustering. According to the cluster results, each image x$_{t}^{i}$ is assigned a pseudo label ${\widetilde{y}}_t^i$. In this way, with the pseudo label of each image in target dataset,  a new training dataset $\left\{(x_t^1,\ {\widetilde{y}}_t^1),\ ...,\ (x_t^i,\ {\widetilde{y}}_t^i),...,(x_t^{N_t},\ {\widetilde{y}}_t^{N_t})\right\}$ is organized.

We then finetune the re-ID model with the new dataset in a supervised manner. Specifically, batch hard triplet loss and cross-entropy loss are used to train the model. Triplet loss is employed with feature vector f$_{t}^{i}$. Cross entropy loss is employed with the person identity classifier, where the global average pooling (GAP) layer and batch normalization (BN) layer are added before it. 
\par Due to the variation of cluster numbers in the training iterations, the newly added person identity classifier layer should be initialized every time executing DBSCAN \cite{ester1996density}. Following the initialization strategy in \cite{zhang2019self}, we exploit the mean features of each cluster as the initial parameters. Specifically, for each cluster \emph{c}, we calculate the mean feature $\bar{f_c}$ by averaging all the embedding features of its elements. The parameters \emph{W$_{c}$} of the classifier layer are initialized as $W_c=\bar{f_c}\ (c=1,2,3,\ ...,C)$, where \emph{C} is cluster number in each iteration. {Finally, a novel adversarial erasing based reconstruction branch is proposed. Specifically, the feature maps of images are used to generate activation maps. Afterward, the coordinates with the largest value are recorded. Then rectangular areas of random size centered on these coordinates are generated as the erase areas. Finally, the erased images are fed into another decoder network for the reconstruction of the original images. A reconstruction loss as $\mathcal{L}_{rec}$ is used. With this strategy, the network can mine the key information from other regions of the erased image, so that the network will not over-trust noisy labels.}

\textbf{\emph{Partial feature branch.}}
In person re-ID task, part-level feature learning usually achieves better performance on discriminative re-ID model learning by mining discriminative body regions. In this paper, we divide the feature maps of person images into two parts horizontally and each part, containing the information about the upper body or lower body, is used to generate a pseudo label of the image separately. Given an unlabeled image x$_{t}^{i}$, the feature map $F_{t}^{i}\in\mathbb{R}^{H\times W\times C}$ is divided into two parts, $F_{t\_up}^{i} \in\mathbb{R}^{\frac{H}{2}\times W\times C}$ and $F_{t\_lower}^{i}\in\mathbb{R}^{\frac{H}{2}\times W\times C}$. Next, the global max pooling (GMP) operation is employed in the sliced feature maps to obtain the feature vectors of different body regions. For every image in target dataset, we extract the feature vectors of upper body and lower body to form two feature vector sets, denoted as $f_{t\_up}=\left\{f_{t\_up}^1,\ ...,f_{t\_up}^k,...,f_{t\_up}^{N_t}\right\}$  and $f_{t\_lower}=\left\{f_{t\_lower}^1,\ ...,f_{t\_lower}^k,...,f_{t\_lower}^{N_t}\right\}$. The unsupervised clustering algorithm is employed on these two vector sets. Therefore, each image x$_{t}^{i}$ is assigned two pseudo labels $\left({\widetilde{y}}_{t\_up}^i, {\widetilde{y}}_{t\_lower}^i\right)$. 

\par Based on the pseudo labels, we cross-finetune the re-ID model in a supervised way with triplet loss, similar to the training strategy of the global branch. Specifically, triplet loss is employed with feature vector \emph{f$_{t\_up}^{i}$} and \emph{f$_{t\_lower}^{i}$}, respectively. 

\textbf{\emph{Global max pooling operation.}}
The procedure of global max pooling and global average pooling is nonparametric. Average pooling calculates the mean of all the pixels within the pooling region while max pooling only considers the maximum response values \cite{zhao2020deep}. For unsupervised clustering, max pooling can produce better performance since the regions with higher response values contains more discriminative information about person images. In the ablation experiments, we will evaluate the effect of global max pooling and global average pooling, and demonstrate the superiority of the former (GMP). 

\textbf{\emph{Loss function.}}
To fine-tune the re-ID model with the pseudo labels generated from global features and part-level features, the full objective function of the two-branch network is formulated as follows, 
\begin{equation}
	\begin{split}
		\mathcal{L}_{all}=\mathcal{L}_{cross}(p_t,\ \widetilde{y_t})+{\lambda_g\mathcal{L}}_{tri}(f_t,\ \widetilde{y_t}) \\
		+\lambda_{up}\mathcal{L}_{tri}(f_{t\_up},\ {\widetilde{y}}_{t\_up})+\lambda_{lower}\mathcal{L}_{tri}(f_{t\_lower},\ {\widetilde{y}}_{t\_lower})
	\end{split}
\end{equation}
where \emph{p$_{t}$} denotes the output sets of the classifier layer in the global branch. $\lambda_g$, $\lambda_{up}$, $\lambda_{lower}$ are hyper-parameters that control the relative importance of global feature triplet loss, upper body feature triplet loss, and lower body feature triplet loss, respectively. In this paper, we use $\lambda_g$=1, $\lambda_{up}$=1, and $\lambda_{lower}$=0.5 in our experiments. Since the upper body of person images usually contains more discriminative regions than the lower body. 

\subsection{Overall algorithm}            
In this section, we describe the overall framework of the proposed method. As shown in Fig. \ref{fig:framework}, our method can be developed in three stages. The first stage learns a pre-trained re-ID baseline model on a labeled source training dataset (Section \ref{subsec:pretrain}). Based on the baseline, the second stage trains the model with images transferred from the source dataset to the target dataset, performing multi-domain image translation (Section \ref{subsec:gan}). In the third stage, we conduct unsupervised clustering and cross-finetune the model with the target training set alternately. Algorithm 1 presents the optimization procedure of our method.

\begin{algorithm}[h]
	\caption{Overall training strategy}
	\label{alg:Overall Training Strategy}
	\begin{algorithmic}[1]
		\REQUIRE ~~\\   
		Labeled source dataset $\left\{X_S,Y_S\right\}$;  \\
		Unlabeled target dataset $\left\{X_T\right\}$;  \\
		Original models $\emptyset_1(\bullet,\theta_{0})$,$\emptyset_2(\bullet,\theta_{0})$;
		\ENSURE ~~\\ 
		Models $\emptyset_1(\bullet,\theta_{t})$ and $\emptyset_2(\bullet,\theta_{t})$;
		\STATE Training baseline model with labeled source dataset $\left\{X_S,Y_S\right\}$;
		\STATE Generating images with target style: input $\left\{X_S,X_T\right\}$;  
		\STATE Training re-ID models with transferred images;
		\STATE Training re-ID models with the two-branch network on unlabeled target dataset $\left\{X_T\right\}$ 
		\FOR{each $n \in [1, num\_epochs]$}
		\STATE  Generate pseudo labels $(y_i^t)$ for each sample $x_i^t$ in $\left\{X_T\right\}$ using global features from another model.
		\STATE  Generate pseudo labels $(y_{i\_up}^t, y_{i\_lower}^t)$ for each sample $x_i^t$ in $\left\{X_T\right\}$ using partial features from another model.
		\FOR{each mini-batch, iterations T}
		\STATE Computing triplet loss with $\left\{f_i^t,y_i^t\right\}$;
		\STATE Computing cross-entropy loss with $\left\{c_i^t,y_i^t\right\}$;
		\STATE Computing triplet loss with $\left\{f_{i\_up}^t,y_{i\_up}^t\right\}$;
		\STATE Computing triplet loss with $\left\{f_{i\_lower}^t,y_{i\_lower}^t\right\}$;
		\STATE Finetune $<\emptyset(\bullet,\ \theta_t)>$ with jointly loss;
		\ENDFOR
		\ENDFOR
	\end{algorithmic}
\end{algorithm}

\section{Experiments}\label{sec:experiment}   
In this section, we evaluate the proposed method on two benchmark datasets: Market-1501 and DukeMTMC-reID, and compare with state-of-the-art unsupervised re-ID methods. 

\subsection{Datasets}  
We conduct experiments on two large-scale benchmark datasets, i.e. Market-1501 \cite{zheng2015scalable} and DukeMTMC-ReID \cite{zheng2017unlabeled,ristani2016performance}. 

\textbf{\emph{Market-1501}} \cite{zheng2015scalable} contains 32,668 images of 1,501 labeled persons from six camera views. Specifically, 12,936 images of 751 identities detected by DPM \cite{felzenszwalb2009object} are used for training. For testing, in total 19,732 images of 750 identities plus some distractors form the gallery set, and 3,368 manually cropped person regions from 750 identities form the query set. 

\textbf{\emph{DukeMTMC-ReID}} \cite{zheng2017unlabeled,ristani2016performance} contains 1,812 identities captured by 8 cameras. There are 16,522 training images, 2,228 query images, and 17,661 gallery images, with 1,404 identities appearing in more than two cameras. Also, similar to the Market-1501, the rest 408 identities are considered as distractors.

\subsection{Implementation Details}\label{subsec:implement details}         
\textbf{\emph{Baseline Model Training. }}
As described in Section \ref{subsec:pretrain}, we first train a baseline model on the source dataset. All training images are resized to $256\times128$ before being fed into the network. For data augmentation, we employ random cropping and flipping. To meet the requirement of hard batch triplet loss, each mini-batch is sampled with randomly selected P = 16 identities and randomly sampled K = 4 images for each identity from the training set, so that the mini-batch size is 64. In the baseline training stage, the margin hyperparameter of triplet loss is set to 0.5 and $\varepsilon$ is set to 0.1 in label smoothing. We use the Adam \cite{kingma2014adam} algorithm with a weight decay 0.0005 to optimize the parameters for 80 epochs. The learning rate is set to 0.00035 and decayed to its 1/10 at the 40th and 70th epochs. During testing, we extract the output of the BN layer as the image descriptor and use the Euclidean distance to compute the similarity between the query and gallery images.

\textbf{\emph{Cross Domain Camera Style Adaptation Model.}}
We follow the same architecture as StarGAN \cite{choi2018stargan}. Specifically, the generator network composes of two convolutional layers with the stride size of two for downsampling, six residual blocks, and two transposed convolutional layers with the stride size of two for upsampling. PatchGANs \cite{isola2017image} are leveraged for the discriminator network. The input images are resized to $256\times128$. We use Adam optimizer with $\beta_1$=0.5 and  $\beta_2$=0.999 for training. The batch size is set to 16. We perform one generator update after five discriminator updates as in \cite{gulrajani2017improved}. In the training stage, we train the model with Market-1501 training dataset and DukeMTMC-reID training dataset simultaneously. In the adaptation stage, for each image in the source set, we generate C style-transferred images (C is the number of cameras in the target set). These C fake images are regarded as containing the same person as the original real images. Then we fine-tune the baseline model with the images transferred from the source dataset to the target dataset.

\textbf{\emph{Unsupervised re-ID model training.}}
For unsupervised training with the target dataset, we train the model in a total of 40 epochs. In each epoch, we perform the unsupervised clustering algorithm and the deep re-ID model training alternately. DBSCAN \cite{ester1996density} clustering method is used to obtain the pseudo labels of each image. Then the new organized dataset is utilized to finetune the re-ID model. For data augmentation, random cropping, random flipping, and random erasing \cite{zhong2020random} are applied. The margin hyperparameter of triplet loss is set to 0.3. We use the Adam with a weight decay of 0.0005 and a learning rate of 0.00035 to optimize the parameters.

\begin{table}[t]
	\caption{Baseline model performance. S: source dataset. T: target dataset.
	}
	\renewcommand\arraystretch{1.2}
	\centering
	\resizebox{0.48\textwidth}{!}{\setlength{\tabcolsep}{1mm}{
			\begin{tabular}{c|cccc|cccc}
				\hline
				\multirow{2}{*}{\diagbox{$S$}{$T$}} & \multicolumn{4}{c|}{Market-1501} & \multicolumn{4}{c}{DukeMTMC-reID}\\
				\cline{2-9}
				& mAP & R1 & R5 & R10 & mAP & R1 & R5 & R10 \\
				\hline
				Market-1501 & 79.6 & 92.6 & 97.2 & 98.3 & 22.3 & 37.6 & 54.2 & 59.9 \\
				DukeMTMC-reID & 24.6 & 53.7 & 69.3 & 74.9 & 69.4	 & 83.3	 & 92.1 & 	94.7 \\
				\hline
	\end{tabular}}}
	\label{table:baseline performance}
\end{table}

\subsection{Experimental Results}     
We conduct cross-domain person re-ID evaluation and compare the results with the state-of-the-art methods, including two hand-crafted features, i.e. Bag-of-Words (BoW) \cite{zheng2015scalable} and local maximal occurrence (LOMO) \cite{liao2015person}, four unsupervised domain adaptation methods, including PTGAN \cite{wei2018person}, SPGAN \cite{deng2018image}, ARN \cite{li2018adaptation} and UDAP \cite{song2020unsupervised}, and six unsupervised methods, including PUL \cite{fan2018unsupervised}, HHL \cite{zhong2018generalizing}, MAR \cite{yu2019unsupervised}, ECN \cite{zhong2019invariance}, SSG \cite{fu2019self}, MMT \cite{ge2020mutual}, DCF \cite{li2021distance}, GLT \cite{zheng2021exploiting} and UNRN \cite{zheng2021group}. Specifically, we use Market-1501 as the source dataset and DukeMTMC-ReID as the target dataset and report the results on DukeMTMC-ReID test set, and vice versa.

\textbf{\emph{Baseline performance.}}   
Here we report the baseline model performance in Table \ref{table:baseline performance}. When trained with a labeled dataset and tested on the same dataset, the baseline model achieves a rank-1 accuracy of 92.6\% and mAP of 79.6\% on Market-1501. However, due to the existence of a domain gap, the performance drops significantly when directly used for another dataset. The rank-1 accuracy declines to 37.6\% when tested on DukeMTMC-reID dataset. 

\textbf{\emph{Comparison with the state-of-the-art methods on Market-1501 Dataset.}}    
Table \ref{table:market result} presents the comparisons when tested on Market-1501 dataset. Compared with four unsupervised domain adaptation approaches, our method outperforms them. The mAP achieves 81.0\%, surpassing the UDA method \cite{song2020unsupervised} by 27.3\%. Compared with other unsupervised methods, which benefit from initializing the model from the labeled source data and learning with unlabeled target data, our method is superior. Comparing with MMT \cite{ge2020mutual}, our results are higher by +7.2\% in rank-1 and +12\% in mAP. 
\begin{table}[t]
	\caption{Comparison with state-of-the-art methods on two datasets. "Duke" and "Market" denote "DukeMTMC-reID" and "Market-1501" respectively.}
	\vspace{-2mm}
	\centering
	\setlength{\tabcolsep}{0.5mm}{\resizebox{0.48\textwidth}{!}{
			\begin{tabular}{c|cccc|cccc}
				\toprule
				\multirow{2}{*}{Method} & \multicolumn{4}{c|}{Duke $\rightarrow$ Market} & \multicolumn{4}{c}{Market $\rightarrow$ Duke}\\
				\cmidrule(r){2-9}
				& rank-1 & rank-5 & rank-10 & mAP & rank-1 & rank-5 & rank-10 & mAP\\
				\midrule
				LOMO~\cite{liao2015person}&27.2&41.6&49.1&8.0 &12.3&21.3&26.6&4.8\\
				Bow~\cite{zheng2015scalable}&35.8&52.4&60.3&14.8 &17.1&28.8&34.9&8.3 \\
				\midrule
				PTGAN~\cite{wei2018person}&38.6&-&66.1&- &27.4&-&50.7&-     \\
				SPGAN~\cite{isola2017image}&51.5&70.1&76.8&22.8 &41.1&56.6&63.0&22.3  \\
				SPGAN+LMP~\cite{isola2017image}&57.7&75.8&82.4&26.7 &46.4&62.3&68.0&26.2   \\
				ARN~\cite{li2018adaptation}&70.3&80.4&86.3&39.4 &60.2&73.9&79.5&33.4    \\
				UDAP~\cite{song2020unsupervised}&75.8&89.5&93.2&53.7  &68.4&80.1&83.5&49.0   \\
				\midrule
				PUL~\cite{fan2018unsupervised}&45.5&60.7&66.7&20.5 &30.0&43.4&48.5&16.4    \\
				HHL~\cite{zhong2018generalizing}&62.2&78.8&84.0&31.4 &46.9&61.0&66.7&27.2     \\ 
				MAR~\cite{yu2019unsupervised}&67.7&81.9&-&40.0   &67.1&79.8&-&48.0          \\
				ECN~\cite{zhong2019invariance}&75.1&87.6&91.6&43.0  &63.3&75.8&80.4&40.4          \\
				SSG~\cite{fu2019self}&80.0&90.0&92.4&58.3 &73.0&80.6&83.2&53.4          \\
				MMT~\cite{ge2020mutual}&86.8&94.6&96.9&69.0  &78.0&88.8&92.5&65.1         \\
				DCF~\cite{li2021distance}&86.1   & 94.2     & 96.0    &  67.6   & 75.8   & 86.5   & 89.4     &  58.3     \\
				GLT~\cite{zheng2021exploiting}& 92.2      & 96.5      &97.8     & 79.5   &82.0    &90.2    &92.8      &69.2       \\
				UNRN~\cite{zheng2021group}& 91.9      &96.1      & 97.8     & 78.1    &82.0    &90.7    &93.5     &69.1       \\
				\midrule
				CMFC(Ours)&94.0&97.1&98.3&81.0 &83.2&91.6&94.0&71.2    \\
				\bottomrule
	\end{tabular}}}
	\label{table:market result}
\end{table}

\textbf{\emph{Comparison with the state-of-the-art methods on DukeMTMC-ReID Dataset.}}   
A similar improvement can be observed in DukeMTMC-reID dataset. As shown in Table \ref{table:market result}, our method achieves rank-1 accuracy = 83.2\% and mAP = 71.2\%, outperforming all the competing UDA methods. For example, comparing with UDAP \cite{song2020unsupervised}, our results are higher by 14.8\% in rank-1 accuracy and 22.2\% in mAP. When compared with unsupervised methods, our method is still superior to most existing methods. Compared to MMT, our method also shows a great advantage in accuracy.

\subsection{Ablation study}    
We also conduct extensive ablation studies to analyze the effectiveness of each component of the proposed method.

\begin {figure}[b]
\centering
\includegraphics[width=0.8\linewidth]{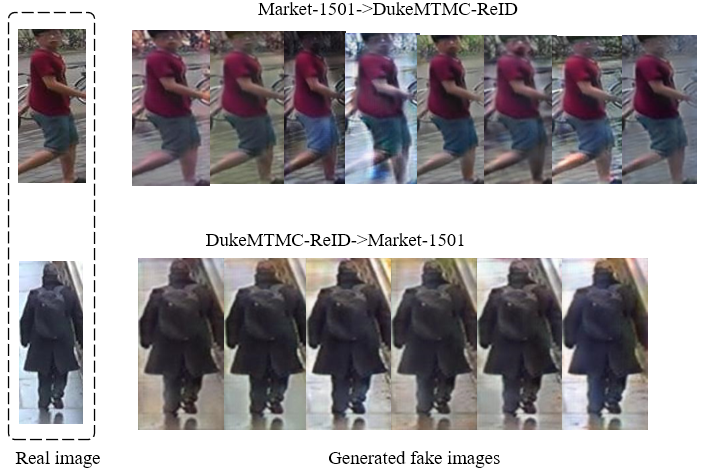}
\caption{Generated samples by cross-domain camera style adaptation module.}
\label{fig:generated samples}
\end{figure}

\textbf{\emph{Effectiveness of Cross Domain Camera Style Adaptation Module.}}  
In the paper, we first train a baseline model on the source dataset. Then we try to reduce the distribution discrepancy at the image level by training a multi-domain image-image generator and transferring images from the source dataset to the target dataset. Specifically, for each image in Market-1501 dataset, we generate eight fake images and assign the same label as the original image. For each image in DukeMTMC-reID dataset, we generate six fake images. Fig. \ref{fig:generated samples} shows generated samples. In this way, we can train the model in a supervised way with the generated images. For example, given Market-1501 as source dataset and DukeMTMC-reID as target dataset, each image in Market-1501 is transferred to eight camera style in DukeMTMC-reID. The generated images with assigned labels are utilized for supervised re-ID training. 
\begin{table}[t]
	\caption{Effectiveness of cross-domain camera style adaptation module. "Duke" and "Market" denote "DukeMTMC-reID" and "Market-1501" respectively.}
	\vspace{-2mm}
	\centering
	\setlength{\tabcolsep}{1mm}{\resizebox{0.48\textwidth}{!}{
			\begin{tabular}{c|cccc|cccc}
				\toprule
				\multirow{2}{*}{Method} & \multicolumn{4}{c|}{Market $\rightarrow$ Duke} & \multicolumn{4}{c}{Duke $\rightarrow$ Market} \\
				\cmidrule(r){2-9}
				& rank-1 & rank-5 & rank-10 & mAP  & rank-1 & rank-5 & rank-10 & mAP \\
				\hline
				\emph{Basel.}&37.6&54.2&59.9&22.3 &53.7&69.3&74.9&24.6\\
				\emph{CDCSA}&53.8&67.8&72.2&31.4  &66.8&82.1&87.9&34.0\\
				\bottomrule
	\end{tabular}}}
	\label{table:ablation cross domain}
\end{table}

We report the results of re-ID model with the transferred images in Table \ref{table:ablation cross domain}. Specifically, when training with the images transferred from Market-1501 to DukeMTMC-ReID, the model achieves rank-1 = 53.8\% and mAP = 31.4\% on DukeMTMC-ReID test set. Compared to PTGAN \cite{wei2018person}, SPGAN \cite{deng2018image} (see Table \ref{table:market result}), our generation models have better results, which also validates that viewing each camera as an individual domain results in better generation quality. Similarly, when training with the images generated from DukeMTMC-ReID, the model achieves rank-1 accuracy = 66.8\% and mAP = 34.0\%, surpassing PTGAN and SPGAN. We further conduct experiments to validate the effectiveness of different loss functions. Specifically, when training with the images generated from original StarGAN with adversarial loss, domain classification loss, and reconstruction loss, the model achieves rank-1 accuracy = 53.4\% and mAP = 30.8\% on DukeMTMC-ReID dataset(see Table \ref{table:ablation cross domain loss item}). Finally, adding identity mapping loss and person identity preserve loss to restrain the training process, our generation models have better results.
\begin{table}
\caption{Effectiveness of different loss functions.}
\vspace{-4mm}
\centering
\setlength{\tabcolsep}{0.8mm}{
	\begin{tabular}{c|cccc}
		\toprule
		\multirow{2}{*}{Method} & \multicolumn{4}{c}{Market-1501 $\rightarrow$ DukeMTMC-reID} \\
		\cmidrule(r){2-5}
		& rank-1 & rank-5 & rank-10 & mAP \\
		\hline
		StarGAN&53.4&67.3&71.8&30.8  \\
		StarGAN+$\mathcal{L}_{idt}$&53.6&67.4&72.8&31.2  \\
		StarGAN+$\mathcal{L}_{pid}$&53.0&67.9&73.0&31.0   \\ 
		StarGAN+$\mathcal{L}_{idt}+\mathcal{L}_{pid}$&53.8&67.8&72.2&31.4 \\
		\bottomrule
\end{tabular}}
\vspace{-1mm}
\label{table:ablation cross domain loss item}
\vspace{-2mm}
\end{table}

\textbf{\emph{Effectiveness of CMFC framework.}}    
To verify the effectiveness of our two-branch network and the collaborative pseudo re-labeling strategy for the target domain, we further conduct experiments over baseline. The results are shown in Table \ref{table:ablation mfc}. Specifically, the model achieves rank-1 accuracy = 83.2\% and mAP = 71.2\% when tested on DukeMTMC-reID dataset. The model achieves 81.0\% and 94.0\% on mAP and rank-1 accuracy when tested on Market-1501. Both the proposed two-branch framework and the collaborative re-labeling strategy boost the overall performance. 

\begin {figure}[b]
\vspace{-2mm}
\centering
\includegraphics[width=1.0\linewidth]{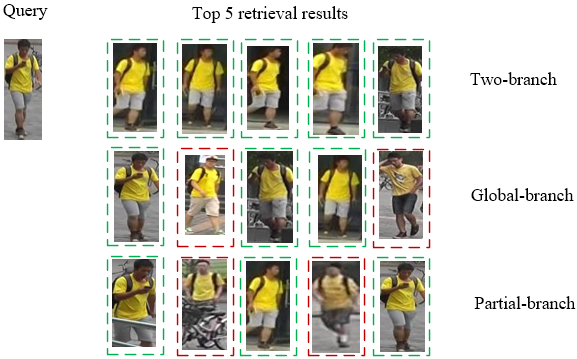}
\caption{Visual retrieval results on Market1501. The green rectangle represents correct matches, and the red dash rectangle represents false matches.}
\label{fig:vis retrieval results}
\vspace{-4mm}
\end{figure}

Specifically, to focus on the contribution of different branches, the ablation studies are evaluated over baseline. Firstly, compared with the baseline model, global and partial feature branches consistently improve the performance, indicating that mining the potential similarity on the target domain is beneficial to discriminative feature learning. The global feature branch obtains higher results than the partial feature branch. We believe that it is because the global feature leads to better clustering quality. Moreover, the partial feature branch can strengthen the representation by mining more information on different body regions. The final two-branch results demonstrate the effectiveness of the combination of global and partial branches. A comparison of visual retrieval results on the Market-1501 dataset between the overall two-branch framework and different branches is shown in Fig. \ref{fig:vis retrieval results}. It can be seen that the two-branch framework achieves better results than both global and partial branches. Besides, we can find that although the cross-domain camera style adaptation module provides a higher baseline, the results of the two-branch network yield a marginal performance increase. From this result, we can find that mining identity similarities on the target domain exhibits more potential.

\begin{table}
\caption{Effectiveness of two-branch framework. "Duke" and "Market" denote "DukeMTMC-reID" and "Market-1501" respectively. "Re-labeling(G/P)" denotes training with only the global or partial branch respectively. "Two-branch" is trained without collaborative re-labeling.}
\vspace{-4mm}
\centering
\setlength{\tabcolsep}{1mm}{\resizebox{0.49\textwidth}{!}{
		\begin{tabular}{c|cccc|cccc}
			\toprule
			\multirow{2}{*}{Method} & \multicolumn{4}{c|}{Duke $\rightarrow$ Market} & \multicolumn{4}{c}{Market $\rightarrow$ Duke} \\
			\cmidrule(r){2-9}
			& rank-1 & rank-5 & rank-10 & mAP  & rank-1 & rank-5 & rank-10 & mAP \\
			\hline
			Basel. &53.7&69.3&74.9&24.6 &37.6&54.2&59.9&22.3\\
			Re-labeling(G) &86.9&93.1&95.2&74.3 &78.2&87.6&90.0&66.2 \\
			Re-labeling(P) &78.2&85.3&88.2&61.4 &69.2&81.6&84.0&58.2 \\
			Two-branch &88.0&94.5&96.3&76.0 &75.1&86.0&89.5&64.9 \\
			CMFC &94.0&97.1&98.3&81.0 &83.2&91.6&94.0&71.2 \\
			\bottomrule
\end{tabular}}}
\vspace{-5mm}
\label{table:ablation mfc}

\end{table}

\textbf{\emph{Effect of different pooling operation.}}     
We also evaluate the effect of different pooing operations on the two-branch framework. We employ global max pooling (GMP) and global average pooling (GAP) operation on the feature maps to obtain feature vectors, respectively. As shown in Table \ref{table:ablation pooling settings}, the model achieves rank-1 accuracy = 91.3\% and mAP = 78.0\% on DukeMTMC-reID dataset with GMP, surpassing the results with GAP. Similar improvement can be observed when tested on Market-1501. The superiority of GMP probably lies in that max pooling filters out some detrimental signals and focuses on the high response values of feature maps, which benefits the discriminative feature extraction of pedestrian images.
\begin{table}
\caption{Influence of different pooling operations. "Duke" and "Market" denote "DukeMTMC-reID" and "Market-1501" respectively.}
\vspace{-4mm}
\centering
\setlength{\tabcolsep}{1mm}{\resizebox{0.49\textwidth}{!}{
	\begin{tabular}{c|cccc|cccc}
		\toprule
		\multirow{2}{*}{Method} & \multicolumn{4}{c|}{Duke $\rightarrow$ Market} & \multicolumn{4}{c}{Market $\rightarrow$ Duke}\\
		\cmidrule(r){2-9}
		& rank-1 & rank-5 & rank-10 & mAP & rank-1 & rank-5 & rank-10 & mAP\\
		\hline
		GMP&91.3&93.3&95.3&78.0 &80.7&89.1&92.0&68.4 \\
		GAP&89.6&90.5&93.8&73.2 &75.2&86.0&90.1&64.7  \\
		\bottomrule
\end{tabular}}}
\label{table:ablation pooling settings}
\vspace{-4mm}
\end{table}

\textbf{\emph{Influences of hyper-parameters.}}     
The weight of different loss items in Eq.13 is a key hyperparameter affecting the performance of feature representation learning. $\lambda_g$, $\lambda_{up}$, and $\lambda_{lower}$ control the relative importance of the whole body, upper body, and lower body similarity constraint. In the experiments, we set $\lambda_g$=1, $\lambda_{up}$=1 and change the value of $\lambda_{lower}$. $\lambda_{lower}$ is set to 0.5 and 1 respectively, and the evaluation results are shown in \ref{table:ablation hyperparameter settings}. Specifically, $\lambda_{lower}$=0.5 yields better accuracy. A possible explanation is that the upper body contains more discriminative information about person images than the lower body. However, it still contributes to the learning of features on the target domain.
\begin{table}
\caption{Influence of $\lambda_{lower}$. "Duke" and "Market" denote "DukeMTMC-reID" and "Market-1501" respectively.}
\centering
\setlength{\tabcolsep}{1mm}{\resizebox{0.49\textwidth}{!}{
	\begin{tabular}{c|ccccc|cccc}
		\toprule
		\multirow{2}{*}{Method}& & \multicolumn{4}{c|}{Duke $\rightarrow$ Market} & \multicolumn{4}{c}{Market $\rightarrow$ Duke}\\
		\cmidrule(r){2-10}
		& & rank-1 & rank-5 & rank-10 & mAP & rank-1 & rank-5 & rank-10 & mAP\\
		\hline
		Two-branch&1                 &92.7&95.9&97.2&79.3 &81.5&90.1&92.2&69.3  \\
		Two-branch&0.5               &94.0&97.1&98.3&81.0 &83.2&91.6&94.0&71.2 \\
		\hline 
		partial-branch&1             &75.4& 81.2&87.5&63.4     &69.6&71.2&84.1&58.5 \\
		partial-branch&0.5           &81.3&86.9&91.3&70.1      &74.2&78.0&86.0&62.3 \\
		\bottomrule
\end{tabular}}}
\label{table:ablation hyperparameter settings}
\vspace{-4mm}
\end{table}

\section{Conclusion}\label{sec:conclusion}             
This paper focuses on unsupervised person re-ID. Specifically, we perform unsupervised domain adaptation on labeled source training images and unsupervised person re-ID on unlabeled target training images. Besides, person identity preserve loss constraint and identity mapping loss constraint are utilized to change the style of images and preserve the identity simultaneously. Moreover, we propose a novel collaborative multiple feature clustering framework (CMFC) for learning representations on the target dataset: global feature guided training branch and partial feature guided training branch. Extensive quantitative experiments validate that learning the potential data similarities on the target domain indeed improves the discriminative representation ability of the person re-ID model. Our method significantly achieves state-of-the-art performance under unsupervised re-ID settings on both two datasets.

\bibliographystyle{ACM-Reference-Format}

\bibliography{sample-base}

\end{document}